\documentclass[a4paper]{article}

\usepackage{INTERSPEECH2015}

\usepackage{graphicx}
\usepackage{amssymb,amsmath,bm}
\usepackage{textcomp}

\usepackage{amssymb,epsfig}
\usepackage{comment,enumitem}
\usepackage{multirow}
\usepackage{comment}
\usepackage[table,xcdraw]{xcolor}
\usepackage{xfrac}
\usepackage{hyperref}

\ninept

\title{Data--selective Transfer Learning for Multi--Domain Speech Recognition}

\makeatletter
\def\name#1{\gdef\@name{#1\\}}
\makeatother \name{\em Mortaza Doulaty, Oscar Saz, Thomas Hain 
	}

\address{Speech and Hearing Group, University of Sheffield, Sheffield, UK \\
  {\small \tt \{mortaza.doulaty, o.saztorralba, t.hain\}@sheffield.ac.uk}
}

\begin{document}

\maketitle

\begin{abstract}

Negative transfer in training of acoustic models for automatic speech recognition has been reported in several contexts such as domain change or speaker characteristics. This paper proposes a novel technique to overcome negative transfer by efficient selection of speech data for acoustic model training. Here data is chosen on relevance for a specific target. A submodular function based on likelihood ratios is used to determine how acoustically similar each training utterance is to a target test set. The approach is evaluated on a wide--domain data set, covering speech from radio and TV broadcasts, telephone conversations, meetings, lectures and read speech. Experiments demonstrate that the proposed technique both finds relevant data and limits negative transfer. Results on a 6--hour test set show a relative improvement of 4\% with data selection over using all data in PLP based models, and 2\% with DNN features.

\end{abstract}
\noindent{\bf Index Terms}: data selection, transfer learning, negative transfer, speech recognition

\section{Introduction} \label{sec:introduction}

As Automatic Speech Recognition (ASR) systems improve their accuracy, new applications and domains 
become the target of research. Automatic transcription of speech with unknown origin is a 
challenging task, which is related to access to so--called ``found data'', such as media and 
historical audio archives. For this to be feasible, ASR has to produce an accurate output for 
whichever the conditions contained in the target data (e.g. interviews, distant recordings, telephone 
conversations, etc). Training acoustic models for an unknown domain, e.g. YouTube recordings, can 
be infeasible if the origin of the target speech can not be properly assessed, and the loss of 
accuracy can be large due to wrong modelling decisions. Another option is to train an acoustic model on 
a large amount of data from multiple domains, although this is not guaranteed to give the most 
optimal results.

Maximum Likelihood Estimation (MLE) of Gaussian Mixture Model (GMM) parameters of a Hidden Markov 
Model (HMM) is still a standard approach to train acoustic models in ASR, either with 
perceptually--based features like Perceptual Linear Prediction (PLP) features \cite{hermansky1990perceptual},
or with Deep Neural Network (DNN) based features \cite{grezl2007probabilistic} in tandem configuration. 
However, MLE has two well--known requirements: first, model correctness is assumed; and second
the amount of training data is required to be infinite \cite{huang2001spoken}. None of the 
above are valid in standard situations in ASR, although systems are sometimes trained with 
many years of speech data (e.g \cite{kapralova2014bigdata}). However, adding more data does not 
guarantee that the performance of the system will improve, and even if it does, the gains
become smaller and smaller \cite{wei2014unsupervised}. A further effect, negative transfer, is
found in several examples,
which indicates that knowledge acquired for a task can have a negative 
performance effect in another task \cite{rosenstein2005transfer}. 
As a result, being able to select informative training data remains an important task.

This paper studies positive and negative transfer in ASR in a multi--domain scenario.
The work proposes to use submodular functions based on acoustic similarity between the target test 
set and training data, in which positive transfer will be exploited to improve performance across 
domains, while reducing  the impact of negative transfer  at the same time. Submodular functions 
have been successfully used before to select data in semisupervised training and active 
learning for ASR tasks \cite{lin2009select, wei2014submodular}. However, here we show that these 
can also be used to select acoustically matching data in an un--supervised manner. 

This paper is structured as follows: Section \ref{sec:related-work} provides a review of data selection techniques for ASR, and Section 
\ref{sec:current-approach} introduces the proposed approach for data selection.
Section \ref{sec:setup} describes the experimental setup, followed by results and analysis in Section 
\ref{sec:results}. The final Section \ref{sec:conclusion} summarises and concludes the paper. 
\section{Data selection for ASR}\label{sec:related-work}
Data selection for ASR has mostly been studied for minimal representative data selection 
\cite{wei2014unsupervised, wei2014submodular, wu2007data, lin2009select, zhang2006new, nagroski2003search, kapralova2014bigdata, gouvea2011kullback, siohan2013ivector}. Here the objective is, given a large pool of 
training data, to find a subset of data such that a model set trained with that data will achieve 
comparable performance to a model set trained with all the data. This line of work is related to 
active learning, where the aim is to select a subset for manual transcription with the least budget 
\cite{riccardi2003active, tur2003active, settles2010active}, and with unsupervised and semi--supervised learning 
techniques, where the overall objective is to select a subset of the training set with the most reliable available transcripts
\cite{wessel2005unsupervised,lanchantin2013automatic,siohan2014training}.



Two techniques are typically used for selecting data: \textit{uncertainty sampling} \cite{zhu2005semi}, where the 
scores from an existing model are used to choose or reject data; and \textit{query by committee} \cite{seung1992query}, 
where votes of distinctly trained models are used \cite{lin2009select}. For uncertainty sampling two 
types of scores have been explored. Confidence scores are used to select data with the 
most reliable transcriptions, as in semi--supervised training \cite{wessel2005unsupervised, 
	kapralova2014bigdata}, or to select data for manual transcription in active learning 
\cite{tur2003active, riccardi2003active}. Entropy--based methods aim to pick data that, for instance, fits a 
uniform distribution of target units (phonemes, words, etc), resulting in maximum entropy 
\cite{lin2009select, zhang2006new,wu2007data} or having a similar distribution to a target set \cite{gouvea2011kullback,siohan2013ivector, siohan2014training}.


The use of submodular functions has been proposed to tackle the effect of the diminishing returns, 
when adding more data to a training set \cite{wei2014submodular, wei2014unsupervised, 
	lin2009select}. A submodular function is defined as any function 
$f:2^{\Omega}\rightarrow\mathbb{R}$ that fulfils
\begin{equation}
\label{eq:submod}
f(S)+f(T) \geq f(S \cup T)+f(S \cap T), \forall S,T \subseteq \Omega
\end{equation}

With submodular functions the problem of data selection turns into a submodular maximisation 
problem, where the objective is to find a subset $S$ from the complete training set $\Omega$ so 
that any new subset $T$ added to $S$ will not increase the value of the submodular function $f$:

\begin{equation}
\label{eq:submax}
\underset{S \subseteq \Omega}{\operatorname{argmax}} \{ f(S) | f(S \cup T) < f(S), \; T \subseteq \Omega \: \backslash \: S \}
\end{equation}

Finding $S$ is an NP--hard problem \cite{krause2012submodular, wei2014submodular} and greedy solutions are proposed where 
the subset $S$ is increased iteratively by the item $s\in\Omega$ that maximises the value of $f$ 
when added to $S$ as
in Equation \ref{eq:greedy}.

\begin{equation}
\label{eq:greedy}
s=\underset{s \in \Omega \backslash S}{\operatorname{argmax}} \{f(S \cup \{s\})\}
\end{equation}
The set $S$ is obtained when either the optimal $S$ is found ($f(S)>f(S\cup\{s\})$), or a budget 
$N$ is reached ($|S| \le N$).

If the function $f$ is a normalised monotone submodular function, then the simple greedy algorithm provides a good approximation of the optimal solution \cite{nemhauser1978analysis, krause2012submodular, lin2009select}

Several functions $f$ can be found in the literature to perform data selection for ASR tasks, 
including facility location functions, saturated coverage functions 
\cite{wei2013using,wei2014submodular}, diversity reward functions \cite{wei2014unsupervised} or 
graph cut functions \cite{lin2009select}.
\section{Likelihood ratio data selection}\label{sec:current-approach}
To decide whether data bears resemblance to a training set, one can opt for a classification 
approach that identifies an item to be suitable or not. Here we make use of the Likelihood Ratio 
(LR) between a GMM trained on the target data ($\Theta_{tgt}$), and a GMM trained on the complete 
training set ($\Theta_{\Omega}$). The total LR of an utterance in the training set 
$LR(\mathcal{O}), \mathcal{O}\in\Omega$ of length $T$ frames is defined as the geometric mean of the 
frame--based LR values of the target data model $\Theta_{tgt}$ and the background model 
$\Theta_{\Omega}$, assuming frame independence.


\begin{equation}
\label{eq:llr}
LR(\mathcal{O})=\frac{1}{T} \sum_{t=1}^{T} \frac{p(\mathcal{O}_t | \Theta_{tgt})}{p(\mathcal{O}_t | \Theta_{\Omega})} 
\end{equation}

One can define a modular function \cite{krause2012submodular} based on the 
accumulated LRs of all utterances included in a subset $S\subseteq\Omega$
in the following form:
\begin{equation}
\label{eq:subllr}
f_{LR}(S)=\sum_{\mathcal{O}\in S}\Big( LR(\mathcal{O}) \Big).
\end{equation}

Modular functions are a special case of submodular functions \cite{krause2012submodular} where the 
greater than or equal sign in Equation \ref{eq:submod} changes to the equal sign. This way, the 
proposed function $f_{LR}$ is submodular as well. And since all of the values for $LR$ are 
non--negative, and therefore any sum of these numbers, as constituted by the function $f$,
the function is necessarily monotonic with expanding sets 
($A \subseteq B \subseteq \Omega, f(A) \le f(B)$). If a submodular function is non--decreasing and 
normalised ($ f(\emptyset)=0 $), then the greedy solution obtained by Equation \ref{eq:greedy} is no 
worse than the optimal value by a constant fraction ($1-\sfrac{1}{e}$) \cite{nemhauser1978analysis}. 
Thus the subset $S$ (greedy solution) can be used as the training set. The stopping  criterion for 
adding more data to this subset $S$ is based on a ``budget'', in the form of a maximum amount of 
hours of speech to be used.

\section{Experimental setup}\label{sec:setup}
To evaluate the proposed approach in a multi--domain ASR task, a data set 
combining 6 different types of data was chosen from the following sources:

\begin{itemize}[noitemsep]
	\item Radio (RD): BBC Radio4 broadcasts on February 2009.
	\item Television (TV): Broadcasts from BBC on May 2008.
	\item Telephone speech (CT): From the Fisher corpus\footnote{All of the telephone speech data was up--sampled to 16 kHz to match the sampling rate of the rest of the data.} \cite{cieri2004fisher}.
	\item Meetings (MT): From AMI \cite{carletta2006ami} and ICSI \cite{janin2003icsi} corpora.
	\item Lectures (TK): From TedTalks \cite{USFD2014IWSLT}.
	\item Read speech (RS): From the WSJCAM0 corpus \cite{robinson1995wsjcam0}.
\end{itemize}

A subset of 10h from each domain was selected to form the training set (60h in total), and 1h from 
each domain was used for testing (6h in total). The selection of the domains aims to cover the most 
common and distinctive types of audio recordings used in ASR tasks.

Two types of acoustic features were used: first, 13 PLP features  plus 
first and second derivatives for a total of 39--dimensional feature vectors; and second, a 
65--dimensional feature vector concatenating the 39 PLP features and 26 bottleneck (BN) features 
extracted from a 4--hidden--layer DNN trained on the full 60 hours of data. 
31 adjacent frames (15 frames to the left and 15 frames to the right) of 23 dimensional log Mel filter bank features were concatenated to form a 713--dimensional super vector; Discrete Cosine Transform (DCT) was applied to this super vector to de--correlate and compress it to 368 dimensions and then fed into the neural network. The network was trained on 4,000 triphone state targets and the 26 dimensional bottleneck layer was placed before the output layer. The objective function used for training was frame--level cross--entropy and the optimisation was performed with stochastic gradient descent using the backpropagation algorithm. DNN training was performed with the TNet toolkit \cite{vesely2010tnet} and more details can be found at \cite{liu2014using}.
For both types of features, MLE--based GMM--HMM models were trained using HTK \cite{young2006htk}
with 5--state crossword triphones and 16 gaussians per state. The language model was based on a 
50,000--word vocabulary and was trained by combination of component language models for each of the 6 domains. 
The interpolation weights were tuned using an independent development set.

\subsection{Baseline results}
Table \ref{tab:baseline} presents results using both types of acoustic features. These results show 
the large variety in performance among domains, from 17--18\% for read speech and radio 
broadcasts to 51\% for television broadcasts. The use of DNN front--ends provides a 25\% relative 
improvement in performance against PLP features; which is consistent across domains and follows 
results previously seen in the literature \cite{yan2013scalable}.


\begin{table}[h]
	\centering
	\caption{WER (\%) of models trained on full set}
	\label{tab:baseline}
	\tabcolsep=0.15cm	
	\begin{tabular}{| c | c | c | c | c | c | c | c |}
		\hline
		Features & RD & TV & CT & MT & TK & RS & Total \\ \hline\hline
		PLP & 18.4 & 51.1 & 46.6 & 44.0 & 34.1 & 17.3 & \textbf{36.0} \\
		PLP+BN & 13.3 & 42.0 & 33.5 & 32.2 & 23.5 & 13.0 & \textbf{26.8} \\
		\hline
	\end{tabular}
\end{table}

\section{Results}\label{sec:results}
An initial set of experiments was conducted to identify and measure negative transfer in ASR 
tasks, and an evaluation of the proposed data selection technique was performed.


\begin{figure}[t]
	\centerline{\epsfig{figure=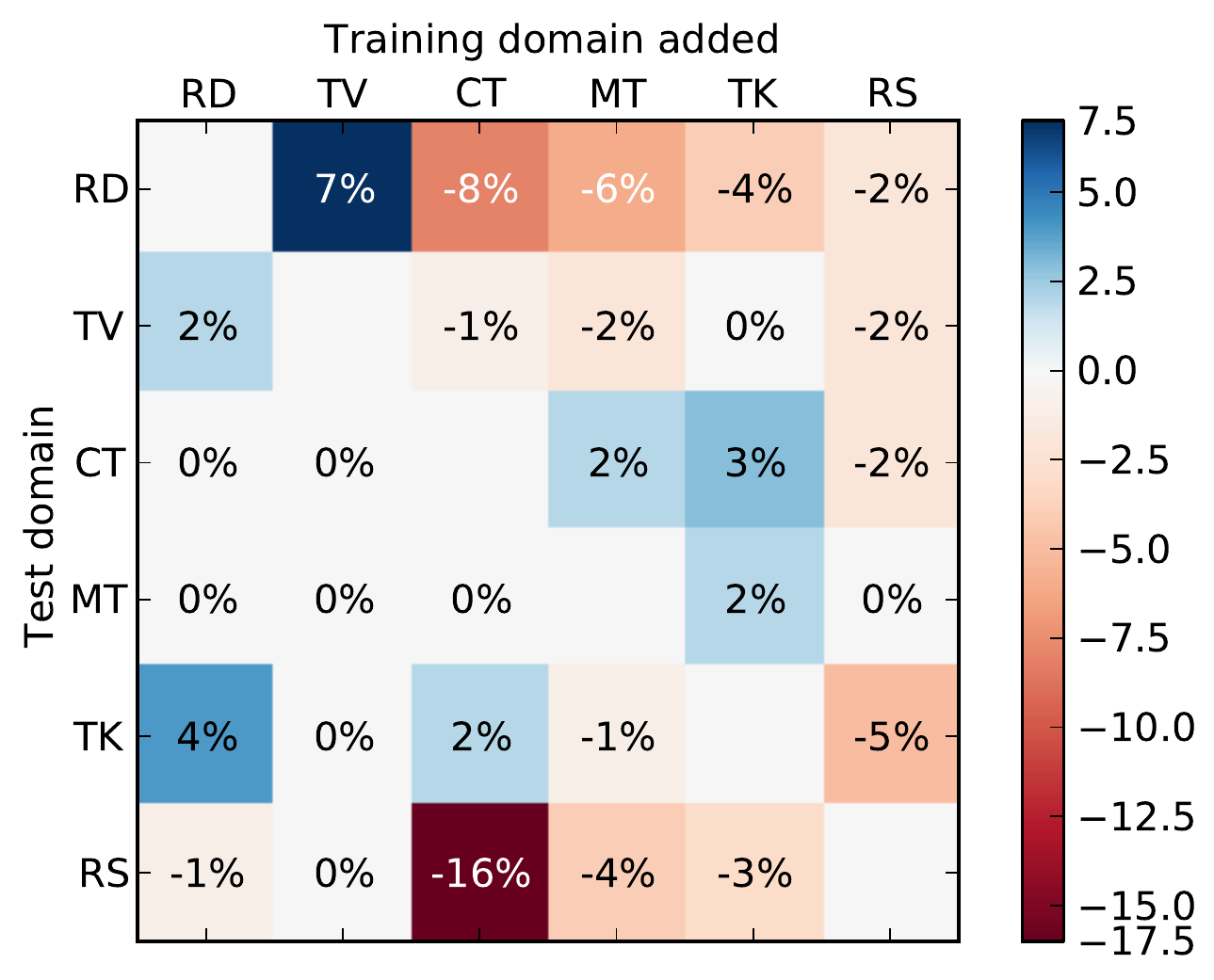,width=80mm}}
	\vspace{-2mm}
	\caption{{\it Relative WER improvement by adding cross--domain data to in--domain models}}  
	\label{fig:transfer}
\end{figure}

\subsection{Evaluation of negative transfer}
Six different domain--dependent MLE models were trained from the 10 hours of training data for each 
domain (in all of the experiments PLP features were used, unless stated otherwise). Each of these 
models was then used to decode the complete test set. The results in Table \ref{tab:domain} show 
that in--domain results (when the train and test data match based on manually labelled domains) are not greatly different from those obtained with a model trained on 
60--hour training set. Instead, cross--domain scores 
(train and test are mismatched) result in considerable performance decreases everywhere.

\begin{table}[h]
	\centering
	\caption{WER (\%) with domain specific acoustic models using PLP features)}
	\label{tab:domain}
	\tabcolsep=0.14cm	
	\begin{tabular}{| c | c | c | c | c | c | c | c |}
		\hline
		Domain & RD & TV & CT & MT & TK & RS & Total \\
		\hline\hline
		RD & \textbf{19.1} & 55.1 & 72.1 & 57.2 & 50.7 & 24.9 & {47.8}   \\
		TV  & 26.5 & \textbf{52.9} & 77.3 & 63.8 & 52.1 & 35.2 & {52.5}\\
		CT & 82.3 & 90.1 & \textbf{44.4} & 71.9 & 67.9 & 86.6 & {72.6}  \\
		MT & 44.9 & 72.3 & 69.2 & \textbf{44.0} & 51.1 & 41.1 & {54.7}\\
		TK & 39.8 & 62.8 & 69.3 & 56.1 & \textbf{35.1} & 55.4 & {53.6} \\
		RS & 29.9 & 66.2 & 84.1 & 67.2 & 68.9 & \textbf{16.9} & {57.4}\\
		\hline    
	\end{tabular}
\end{table}

A second set of experiments was performed with models trained on 20 hours of data, using data from 
every possible pair of domains, for a total of 30 new acoustic models. Figure \ref{fig:transfer} 
shows the results in terms of relative improvement and degradation over the results of the 10--hour in--domain 
models. The rows of Figure \ref{fig:transfer} represent the testing domain and the columns represent 
the domain that was added in training to the data of the domain of the row. Positive values (blue 
squares) mark the existence of positive transfer, such as adding TV data to Radio data (7\% improvement) or adding 
Radio data to Lecture data (4\% improvement). But negative values (red squares) mark negative transfer, like adding 
Telephone data to Read speech (16\% loss) or adding Read speech to Lecture data (5\% loss).

These results showed that positive and negative transfer occurred across domains, possibly due to similarities and differences in speech styles, acoustic channels and background conditions. However a rule--based optimisation of the best model
for each target domain would require a complex and error--prone process. The next experiments aimed to evaluate how an
automatic selection of training could exploit positive transfer, while restricting negative transfer.

\subsection{Data selection based on budget}
The data selection technique proposed in Section \ref{sec:current-approach} was evaluated next. For each of the six target test domains, Gaussian Mixture Models (GMM) with 512 mixtures were trained ($\Theta_{tgt_{1: 6}}$), and 
a background 512--mixture GMM ($\Theta_{\Omega}$) was trained from the complete training set of 60 
hours. These GMMs were used to calculate the LR value for each training utterance 
($LR(\mathcal{O}$)) in order select the training data according to the acoustic similarity.

\begin{figure}[t]
	\centerline{\epsfig{figure=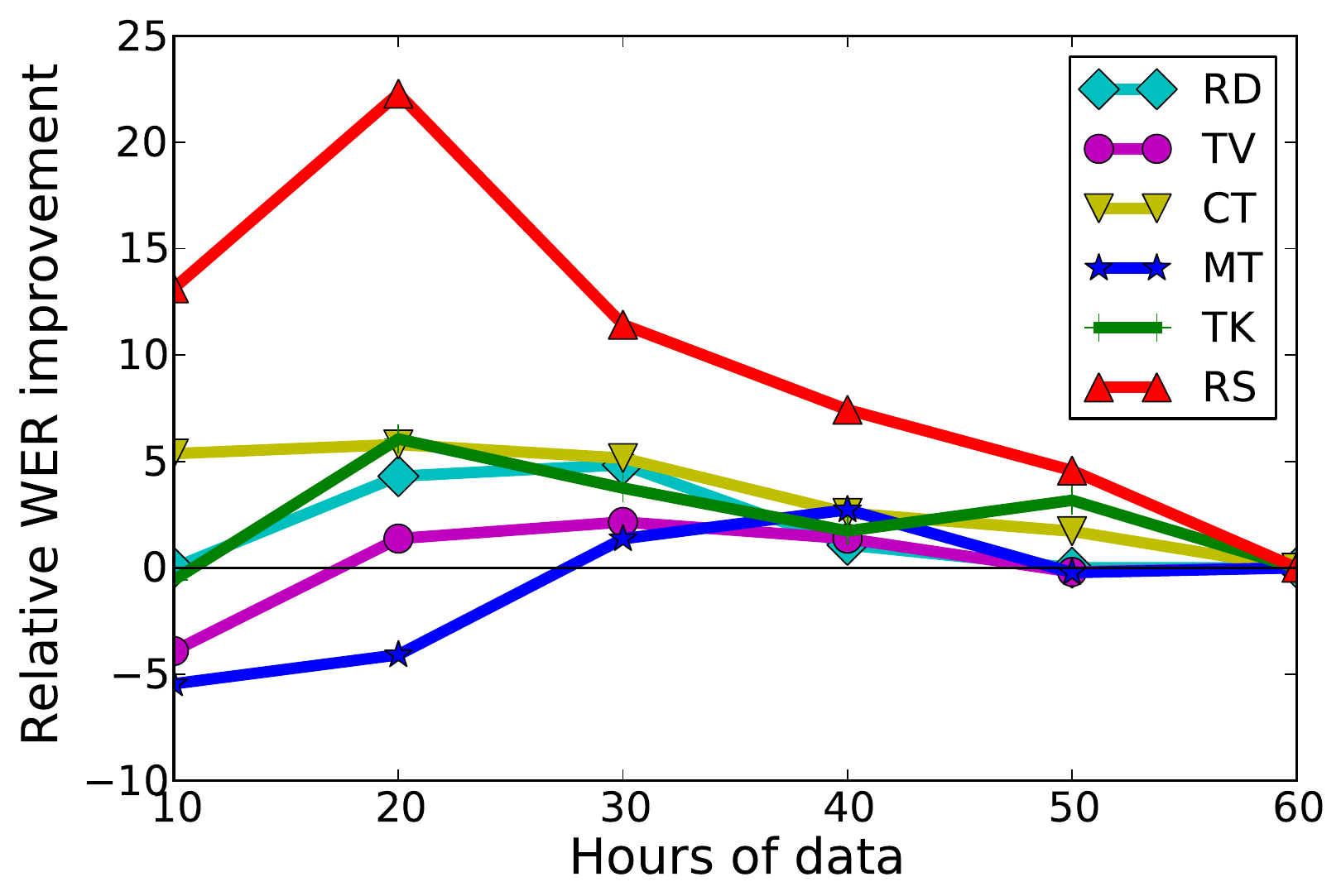,width=80mm}}
	\vspace{-2mm}
	\caption{{\it WER improvement with budget--based data selection}}  
	\label{fig:ll-ratio-selection-wer-rel}
\end{figure}

The first evaluation was performed using data selection based on budget. Five possible budgets of 
10, 20, 30, 40 and 50 hours were designed for each test domain and the respective training data was 
chosen using the $f_{LR}(S)$ submodular function. Figure \ref{fig:ll-ratio-selection-wer-rel} shows 
relative improvement for each domain and budget against the results with the 60--hour model. The 
graphs show that all domains improve performance as the budget increases until a certain limit is 
reached, then negative transfer decreases the performance, converging to the WER achieved with the 
60--hour trained model.

\begin{figure}[t]
	\centerline{\epsfig{figure=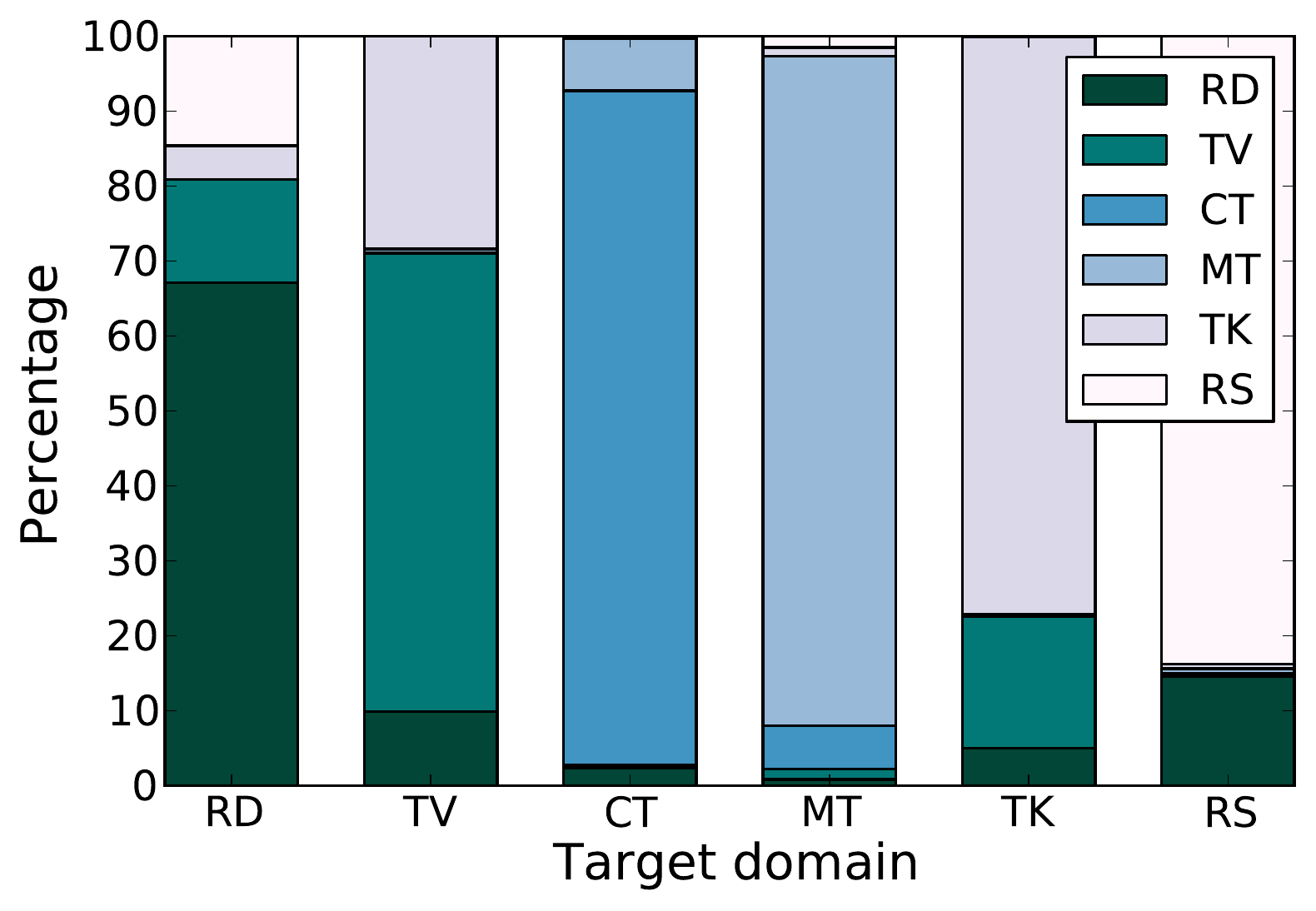,width=80mm}}
	\vspace{-2mm}
	\caption{{\it Types of data selected for a 10--hour budget}}  
	\label{fig:10h-ll-ratio-selection-percentage}
\end{figure}


In order to observe which types of data were selected for each domain with the different budgets, 
Figure \ref{fig:10h-ll-ratio-selection-percentage} presents the percentage of training data selected 
for each test domain with a 10--hour budget. While the majority of the data was chosen from the same 
domain, some cross--domain data was also selected, indicating positive transfer between domains. This 
occurred, for instance, with TV and Read speech data towards Radio data; and Lecture data towards TV data. 





\subsection{Automatic decision on budget}
An issue that can arise with the evaluated budget--based proposal is the fact that a decision on a budget has to be made, and as the results in Figure \ref{fig:ll-ratio-selection-wer-rel} suggest, the optimal budget varies across different domains. A method for deciding a budget for a given target domain was proposed by selecting only utterances whose likelihood--ratio is above a threshold defined as the mean of the highest--weighted mixture of a GMM fitted to the distribution of likelihood ratios. The use of the mixture with the highest weight avoids the influence of outliers in the distribution of the LR values. 

The experiments with an automatic budget decistion were performed for both types of features, 
PLP and PLP+BN. Table \ref{tab:dnn-30h} presents the results for these experiments and compares them 
to the outcome of data selection based on a 30--hour budget, which was the best fixed budget from Figure \ref{fig:ll-ratio-selection-wer-rel}. The results showed that the use of an automatically derived threshold improved the results obtained with a fixed budget for both types of features, indicating that the proposed method could estimate the right amount of data to
select for each target domain.

\begin{table}[h]
	\centering
	\caption{WER(\%) using data selection}
	\label{tab:dnn-30h}
	\tabcolsep=0.14cm	
	\begin{tabular}{|c|c | c | c | c | c | c | c |}
		\hline
		Method & RD & TV & CT & MT & TK & RS & Total \\
		\hline\hline
		\multicolumn{8}{|c|}{PLP features}\\
		\hline\hline
		Budget--30h. & 17.7 & 50.0 & 44.2 & 43.4 & 33.4 & 15.5 & \textbf{34.9} \\
		Auto. Decision    & 17.7 & 49.7 & 44.2 & 43.8 & 32.9 & 15.1 & \textbf{34.7}\\
		\hline\hline
		\multicolumn{8}{|c|}{PLP+BN features}\\
		\hline\hline
		Budget--30h. & 13.0 & 41.5 & 32.6 & 32.1 & 22.5 & 12.1 & \textbf{26.3} \\
		Auto. Decision    & 12.7 & 41.4 & 32.5 & 32.3 & 22.4 & 11.8 & \textbf{26.2}\\
		\hline
		
	\end{tabular}
\end{table}

The amount of 
data selected for each domain is presented in Table \ref{tab:data}. This Table shows how Read speech 
and Conversational Telephone speech are the ones which benefited from a lower amount of training 
data (20 hours or less), while the rest of the domains preferred more data (from 30 to 40 hours). 
These values were consistent with the patterns of positive and negative transfer observed in Figure 
\ref{fig:ll-ratio-selection-wer-rel}.

\begin{table}[h]
	\centering
	\caption{Hours of data selected by automatic budget decision}
	\label{tab:data}
	\tabcolsep=0.14cm	
	\begin{tabular}{|c|c | c | c | c | c | c |}
		\hline
		Domain & RD & TV & CT & MT & TK & RS \\
		\hline\hline
		Hours & 41.2 & 35.8 & 21.9 & 35.6 & 31.4 & 17.1 \\
		\hline
		
	\end{tabular}
\end{table}

\section{Conclusion}\label{sec:conclusion}
In this paper, the effect of positive and negative transfer across widely diverse domains in ASR 
was explored. We confirmed that the use of more data in MLE--based acoustic models does not 
always provide increases in performance. A submodular function based on Likelihood Ratio
was proposed and used to perform an informed and efficient selection of data for different target 
test sets. The evaluation of selection techniques based on budget and on automatic budget decision
has achieved gains of 4\% over a 60--hour MLE model for PLP features and 2\% 
for PLP+BN features.

Previous works have shown that data selection techniques can result in data sets biased towards specific groups of phones or triphones \cite{siohan2014training}. A phonetic analysis
of the data sets given by the likelihood ratio function used in this paper did not show any bias on phones in these data sets. The 60--hour training data used in this work was well balanced phonetically which limited the risk of phonetic biases in the selected data. In situations where the original training data might present less well distributed phonetic content, the proposed function should be complemented by a function that takes into account the resulting phone distribution of the data.

Future work should explore similar data selection techniques for other training criteria besides 
MLE. The presented methods are based on LR and hence well--suited for MLE, but other submodular 
functions will be required to cater for needs given by discriminative objective functions such as 
Minimum Phone Error training. Further work should also investigate data 
selection techniques for datasets larger than the one studied here, and in completely mismatched 
conditions and using different features that better describe the background's acoustic characteristics \cite{saz2014slt}.

The technique presented in this paper can be used for building targeted models for ``found 
speech data''.
The ability of using very diverse data sets to transcribe newly found sets of speech recorded in unknown 
conditions is especially necessary to deal with this type of data. Other tasks, such as the 
automatic transcription of multi--genre media archives might also potentially benefit from the 
advances achieved in this work.

\section{Acknowledgements}\label{sec:acknowledgements}  
This work was supported by the EPSRC Programme Grant EP/I031022/1 Natural Speech Technology (NST).

\section{Data Access Statement}
The speech data used in this paper was obtained from the following sources: Fisher Corpus (LDC catalogue number LDC2004T19), ICSI Meetings corpus (LDC catalogue number LDC2004S02), WSJCAM0 (LDC catalogue number LDC95S24), AMI corpus (DOI number 10.1007/11677482\_3), TedTalks data (freely available as part of the IWSLT evaluations), BBC Radio and TV data (this data was distributed to the NST project's partners with an agreement with BBC R\&D and not publicly available yet).


The specific file lists used for training and testing in the experiments in this paper, as well as result files can be downloaded from \url{http://mini.dcs.shef.ac.uk/publications/papers/is15-doulaty}.
  \eightpt
  \bibliographystyle{IEEEtran}
  \bibliography{references}

\end{document}